\documentclass[a4paper]{article}

\usepackage[english]{babel}
\usepackage[utf8x]{inputenc}
\usepackage[T1]{fontenc}

\usepackage[a4paper,top=3cm,bottom=2cm,left=3cm,right=3cm,marginparwidth=1.75cm]{geometry}

\usepackage{amsmath, amsfonts,amssymb,amsthm}
\usepackage{graphicx}
\usepackage[colorinlistoftodos]{todonotes}
\usepackage[colorlinks=true, allcolors=blue]{hyperref}

\usepackage{lineno,hyperref}
\usepackage{subfigure}
\usepackage{algorithmic}
\usepackage{float}
\usepackage[affil-it]{authblk}
\usepackage{cases}
\usepackage{multirow}

\title{Fish School Search Algorithm for Constrained Optimization}
\author[1]{J.B. Monteiro-Filho*}
\author[1]{I.M.C. Albuquerque}
\author[1]{F.B. Lima Neto}
\affil[1]{Computational intelligence research group - Polytechnical School of Pernambuco Benfica 455, Recife-PE, Brazil}
\affil[*]{Corresponding author: jbmf@ecomp.poli.br}

\begin{document}
\maketitle

\begin{abstract}
In this work we investigate the effectiveness of the application of niching able swarm metaheuristic approaches in order to solve constrained optimization problems. Sub-swarms are used in order to allow the achievement of many feasible regions to be exploited in terms of fitness function. The niching approach employed was wFSS, a version of the Fish School Search algorithm devised specifically to deal with multi-modal search spaces. A base technique referred as wrFSS was conceived and three variations applying different constraint handling procedures were also proposed. Tests were performed in seven problems from CEC 2010 and a comparison with other approaches was carried out. Results show that the search strategy proposed is able to handle some heavily constrained problems and achieve results comparable to the state-of-the-art algorithms. However, we also observed that the local search operator present in wFSS and inherited by wrFSS makes the fitness convergence difficult when the feasible region presents some specific geometrical features.
\end{abstract}

\section{Introduction}

% no \IEEEPARstart

According to Koziel and Michalewicz \cite{Koziel1999}, the general nonlinear programming problem (NLP) consists in finding $\textbf{x}$ such that:\\

$optimize \: f(\textbf{x}), \: \textbf{x}=(x_1,...,x_n) \in \mathbb{R}^n,$\\
\\where $\textbf{x} \in \mathcal{F} \subseteq \mathcal{S}$. Objective function $f$ is defined on the search space $\mathcal{S} \subseteq \mathbb{R}^n$ and the set $\mathcal{F} \subseteq \mathcal{S}$ defines the feasible region.

Search space $\mathcal{S}$ is defined as a rectangle within $\mathbb{R}^n$ and $m \geq 0$ constraints define the feasible space $\mathcal{F}\subseteq\mathcal{S}$:\\

$g_j(\textbf{x}) \leq 0, \: \text{for} \: j=1,...,q,$

and $h_j(\textbf{x}) = 0, \: \text{for} \: j=q+1,...,m.$\\

Equality constraints are commonly relaxed and transformed in inequality constraints as \cite{Mezura-Montes2010}: $|h_j(\textbf{x})| - \delta \leq 0$, where $\delta$ is a very small tolerance value.

Almost all real-world optimization problems are constrained \cite{Jordehi2015}. Hence, many metaheuristic search procedures were conceived for the general NLP. Recent approaches include: Genetic Algorithms \cite{Lin2013, Deb2000, Koziel1999}, Diferential Evolution \cite{Liu2016, Hamza2015, Takahama2010, Brest2009, takahama2009, takahama2006}, Cultural Algorithm \cite{Becerra2016}, Particle Swarm Optimization \cite{Jordehi2015, Campos2013, Bonyadi2013, Liang2010, takahama2005, Hu2002} and Artificial Bee Colony Optimization \cite{Li2014, Brajevic2013, Akay2012, Mezura-Montes2010}.

Regarding the approaches applied in order to tackle constrained search, Mezura-Montes and Coello Coello \cite{Mezura2011} present a simplified taxonomy of the common procedures in literature:

\begin{enumerate}
\item Penalty functions - Includes a penalization term in the objective function due to constraint violation. This is a popular and easy to implement approach but has the drawback of requiring the penalty weights adjustment;
\item Decoders - Consists in mapping the feasible region on search spaces where an unconstrained problem will be solved. The high computational cost required is the main disadvantage in its use;
\item Special operators - Mainly in evolutionary algorithms, operators can be designed in a way to prevent the creation of infeasible individuals;
\item Separation of objective function and constraints - This approach, different from penalty functions, treat objective function and constraint violations in separately. Many procedures can be applied from this approach such as dividing the search process in phases or applying multi-objective techniques.
\end{enumerate}

Fish School Search (FSS) algorithm, presented originally in 2008 in the work of Bastos-Filho and Lima-Neto et al. \cite{Filho2008}, is a population based continuous optimization technique inspired in the behavior of fish schools while looking for food. Each fish in the school represents a solution for a given optimization problem and the algorithm utilizes information of every fish to guide the search process to promising regions in the search space as well as avoiding early convergence in local optima.

Ever since the original version of FSS algorithm was developed, many modifications were performed in order to tackle different issues such as multi-objective optimization \cite{Bastos2015}, multi-solution optimization \cite{Madeiro2011} and binary search \cite{Sargo2014}. Among those, a novel niching and multi-solution version known as wFSS was recently proposed \cite{DeLimaNeto2013}.

To the best of the authors knowledge, the application of FSS in the solution of constrained optimization problems has never been reported before. Hence, in this work, a modification in niching weight based FSS (wFSS) was carried out. The separation of objective function and constraints was applied and the niching feature was used in order for the population to find different feasible regions within the search space to be exploited in terms of fitness value. Moreover, three different mechanisms were applied generating in total four approaches.

This paper is organized as follows: section \ref{FSS} provides an overview of Fish School Search algorithm and its niching version, wFSS. Section \ref{wrFSS} introduces the proposed modifications in order to employ wFSS in constrained optimization problems as well as the variations introduced based on the main procedure conceived. Section \ref{experiments} presents the tests performed and results achieved.

\section{Fish Schooling Inspired Search Procedures}
\label{FSS}
\subsection{Fish School Search Algorithm}

FSS is a population based search algorithm inspired in the behavior of swimming fishes in a school that expands and contracts while looking for food. Each fish $n$-dimensional location represents a possible solution for the optimization problem. The algorithm makes use of weights for all fishes which represent cumulative account on how successful has been the search for each fish in the school.

FSS is composed of feeding and movement operators, the latter being divided into three sub-components, which are:

\begin{enumerate}

\item \textbf{Individual component of the movement:} Every fish in the school performs a local search looking for promising regions in the search space. It is done as represented by (\ref{indMov}):

\begin{equation}
\label{indMov}
\textbf{x}_i(t+1)=\textbf{x}_{i}(t)+\textbf{rand(-1,1)}step_{ind},
\end{equation}

where $\textbf{x}_{i}(t)$ and $\textbf{x}_{i}(t+1)$ represent the position of fish $i$ before and after the individual movement operator, respectively. $\textbf{rand(-1,1)}$ is an uniformly distributed random numbers array with the same dimension as $\textbf{x}_{i}(t)$ and values varying from $-1$ up to $1$. $step_{ind}$ is a parameter that defines the maximum displacement for this movement. The new position $\textbf{x}_i(t+1)$ is only accepted if the fitness of fish $i$ improves with the position change. If it is not the case, $\textbf{x}_{i}(t)$ remains the same and $\textbf{x}_i(t+1)=\textbf{x}_{i}(t)$.

\item \textbf{Collective-instinctive component of the movement: } An average of displacements performed within individual movements is calculated based (\ref{colInst}):

\begin{equation}
\label{colInst}
\textbf{I}=\frac{\sum^{N}_{i=1} \Delta \textbf{x}_{i} \Delta f_{i}}{\sum^{N}_{i=1} \Delta f_{i}}.
\end{equation}

Vector $\textbf{I}$ represents the weighted average of the displacements of each fish. It means that fishes which experienced a higher improvement will attract other fishes into its current position.

After vector $\textbf{I}$ computation, every fish will be encouraged to move according to (\ref{colInstMov}):

\begin{equation}
\label{colInstMov}
\textbf{x}_i(t+1)=\textbf{x}_{i}(t)+\textbf{I}.
\end{equation}

\item \textbf{Collective-volitive component of the movement: } This operator is used in order to regulate exploration/exploitation abilities of the school during the search process. First of all, barycenter $\textbf{B}$ is calculated based on the position $\textbf{x}_{i}$ and weight $W_{i}$ of each fish:

\begin{equation}
\textbf{B}(t)=\frac{\sum^{N}_{i=1} \textbf{x}_{i}(t) W_{i}(t)}{\sum^{N}_{i=1} W_{i}(t)},
\end{equation}	
and then, if total weight given by the sum of weights of all $N$ fishes in the school $\sum^{N}_{i=1} W_{i}$ has increased from last to current iteration, the fishes are attracted to the barycenter according to (\ref{volAttrac}). If the total school weight has not improved, fishes are spread away from the barycenter according to (\ref{volSpread}):

\begin{equation}
\label{volAttrac}
\textbf{x}_i(t+1)=\textbf{x}_{i}(t) - step_{vol} \textbf{rand(0,1)} \frac{\textbf{x}_{i}(t) - \textbf{B}(t)}{distance(\textbf{x}_{i}(t),\textbf{B}(t))},
\end{equation}

\begin{equation}
\label{volSpread}
\textbf{x}_i(t+1)=\textbf{x}_{i}(t) + step_{vol} \textbf{rand(0,1)} \frac{\textbf{x}_{i}(t) - \textbf{B}(t)}{distance(\textbf{x}_{i}(t),\textbf{B}(t))},
\end{equation}

where $step_{vol}$ defines the maximum displacement performed with the use of this operator. $distance(\textbf{x}_{i}(t),\textbf{B}(t))$ is the euclidean distance between fish $i$ position and the school barycenter. $\textbf{rand(0,1)}$ is an uniformly distributed random numbers array with the same dimension as $\textbf{B}$ and values varying from $0$ up to $1$.

\end{enumerate}

Besides movement operators, it was also defined a feeding operator used in order to update the weights of every fish according to (\ref{nfFeed}):

\begin{equation}
\label{nfFeed}
W_{i}(t+1)=W_{i}(t)+\frac{\Delta f_i}{max(| \Delta f_i |)},
\end{equation}
where $W_{i}(t)$ is the weight parameter for fish $i$, $\Delta f_i$ is the fitness variation between the last and new positions and $max(| \Delta f_i |)$ represents the maximum absolute value of fitness variation among all fishes in the school.

$W$ is only allowed to vary from 1 up to $W_{scale}$, which is a user defined attribute. Weights of all fishes are initialized with the value $W_{scale}/2$.

The parameters $step_{ind}$ and $step_{vol}$ decay linearly along with the iterations.

\subsection{Weight-based Fish School Search Algorithm}
Introduced in the work of Lima-Neto and Lacerda \cite{DeLimaNeto2013}, wFSS is a weight based niching version of FSS intended to provide multiple solutions for multi-modal optimization problems. The niching strategy is based on a new operator called Link Formator. This operator is responsible for defining leaders for fishes in order to form sub-schools and works according to the following: each fish $a$ chooses randomly another fish $b$ in the school. If $b$ is heavier than $a$, then $a$ now has a link with $b$ and follows $b$ (i.e. b leads a). Otherwise, nothing happens. However, if $a$ already has a leader $c$ and the weights sum of $a$ followers is higher than $b$ weight, then $a$ stops following $c$ and starts following $b$. In each iteration, if $a$ becomes heavier than its leader, the link will be broken. 

In addition to Link Formator operator inclusion, some modifications were performed in the components of the movement operators in order to emphasize leaders influence on sub-swarms. Thus, the displacement vector $\textbf{I}$ of the collective-instinctive component becomes:
\begin{equation}
\textbf{I}=\frac{\Delta \textbf{x}_{i} \Delta f_{i} + L \Delta \textbf{x}_{l} \Delta f_{l}  }{\Delta f_{i} + L \Delta f_{l}},
\end{equation}
where $L$ is 1 if fish $i$ has a leader and $0$ otherwise. $\Delta \textbf{x}_{l}$ and $\Delta f_{l}$ are the displacement and fitness variation of the leader of fish $i$. Furthermore, the influence of vector $\textbf{I}$ in fishes movements is increased along with iterations. This is represented by $\textbf{x}_i(t+1)=\textbf{x}_{i}(t)+ \rho \textbf{I}$ with $\rho = \frac{current iteration}{It_{max}}$. The collective-volitive component of the movement is also modified in a sense that the barycenter is now calculated for each fish with relation to its leader. If the fish does not have a leader, its barycenter will be its current position. This means:
\begin{equation}
\textbf{B}(t)=\frac{\textbf{x}_{i}(t) w_{i}(t) + L \textbf{x}_{l}(t) w_{l}(t)}{w_{i}(t) + L w_{l}(t)},
\end{equation}

\section{wrFSS}
\label{wrFSS}

Some modifications were proposed in wFSS in order to make the algorithm able to tackle constrained optimization problems. Basically, either fitness values and constraint violation are measured for every fish. In the beginning of each iteration, a decision has to be done in order to define whether fitness function or constraint violation will be used as objective function within current iteration.

The decision of which value to use as objective function was chosen to be done according to the feasible individuals proportion with relation to whole population. This means that, if the current  feasible proportion of the population is higher than an user defined threshold $\sigma$, the search will be performed using fitness as objective function. If that is not the case, constraint violation will be then optimized.

The described procedure was applied to divide the search process in two different phases and to allow the algorithm to: \textit{phase 1} - find many feasible regions; \textit{phase 2} - optimize fitness within feasible regions. The niching feature of wFSS is useful in phase 1 once this feature will make the school able to find many different feasible regions.  Moreover, every once the search changes from phase 1 to phase 2, an increase factor $\tau$ is applied in the steps of either Individual and Collective-volitive movement operators in order to augment the school mobility in the new phase.

The algorithm described will be referred as wrFSS and its pseudocode is:

\begin{algorithmic}[1]
\STATE Initialize user parameters 
\STATE Initialize fishes positions randomly 
\WHILE{Stopping condition is not met}
\STATE Calculate fitness for each fish
\STATE Calculate constraint violation for each fish
\IF{$Feasible \: proportion \geq \sigma$} \STATE {To define fitness as objective function} \ELSE \STATE{To define constraint violation as objective function} \ENDIF
\STATE Run individual movement operator
\STATE Run feeding operator 
\STATE Run collective-instinctive movement operator
\STATE Run collective-volitive movement operator
\ENDWHILE	
\end{algorithmic}

The constraint violation measure applied in wrFSS was the same as in the work of Takahama and Sakai \cite{Takahama2010}:

\begin{equation}
\phi(\textbf{x}) = \sum_{j=1}^{q} \lvert \text{max} \{0,g_j(\textbf{x})\} \rvert ^p + \sum_{j=q+1}^{m} \lvert \{0,h_j(\textbf{(x)})\} \rvert ^p.
\end{equation}

Best fish selection was done using Deb's heuristic \cite{Deb2000}:

\begin{enumerate}
\item Any feasible solution is preferred to any infeasible solution;
\item Among two feasible solutions, the one having better fitness function will be preferred;
\item Among two infeasible solutions, the one having smaller constraint violation is preferred.	
\end{enumerate}

Furthermore, the feeding operator version applied was the same as in the work of Monteiro et al. \cite{monteirob2016}. In this version, feeding becomes a normalization of both fitness and constraint violation values:

\begin{equation}
\label{feedNormalized}
W_i = W_{scale} + (1-W_{scale}) \dfrac{f_i-min(f)}{max(f)-min(f)},
\end{equation}
where $f$ will be constraint violation values within phase 1 and fitness on phase 2. $min(f)$ and $max(f)$ are the minimum and maximum $f$ values found in all the search process.

It is important to highlight that the normalization applied in Equation \ref{feedNormalized} makes $max(f) \Longrightarrow 1$ and $min(f) \Longrightarrow W_{scale}$ once this equation is applied for minimization of both fitness function and constraint violation.

\subsection{wrFSS Variations}

In this Section, some variations of the aforementioned algorithm will be presented applying state-of-the-art constrained optimization approaches within wrFSS. wrFSS variations are:

\begin{enumerate}
\item wrFSSe - Applies the $\epsilon$-method \cite{takahama2005} in Individual component of the movement;
\item wrFSSg - Includes a gradient based local search in either phases of the search process;
\item wrFSSp - Uses a penalized fitness function in phase 2.
\end{enumerate}

wrFSS variations are intended to increase the algorithm performance when tackling challenging problems. Each of the variations specific mechanisms will be better described in next sections.

\subsubsection{wrFSSe}

The $\epsilon$-method \cite{takahamab2005, takahama2009, Takahama2010, Brest2009} defines a comparison procedure taking simultaneously into account constraint violation and fitness value.
Let $f_i$ and $\phi_i$ be the fitness value and constraint violation evaluated at point $\textbf{x}_i$. Thus, $\epsilon$ comparisons $<_{\epsilon}$ and $\leq_{\epsilon}$, with $\epsilon \geq 0$, are defined as:

\begin{numcases}{(f_1, \phi_1) <_{\epsilon} (f_2, \phi_2) \Leftrightarrow}
f_1 < f_2, & if $\phi_1, \phi_2 \leq \epsilon$ \\
f_1 < f_2, & if $\phi_1 = \phi_2$ \\
\phi_1 < \phi_2, & Otherwise,
\end{numcases}

\begin{numcases}{(f_1, \phi_1) \leq_{\epsilon} (f_2, \phi_2) \Leftrightarrow}
f_1 \leq f_2, & if $\phi_1, \phi_2 \leq \epsilon$ \\
f_1 \leq f_2, & if $\phi_1 = \phi_2$ \\
\phi_1 \leq \phi_2, & Otherwise.
\end{numcases}

When $\epsilon \rightarrow \infty$, $\epsilon$-comparison becomes a simple fitness comparison. Further, for $\epsilon = 0$, Deb's heuristic is carried out.

In wrFSSe, $\epsilon$ was chosen to decay along with the iterations in the same way as in the work of Takahama and Sakai \cite{Takahama2010}:

\begin{numcases}{\epsilon(t) =}
\epsilon_{0}\left(1-\dfrac{t}{T_c}\right)^{cp}, & $0 < t < T_c$ \\
0, & $t \geq T_c$,
\end{numcases}
where $t$ is the current iteration and $T_c$ is a percentage of the maximum number of iterations. $cp$ is given by $max\left(cp_{min}, \frac{-5-log \epsilon(0)}{log0.05}\right)$, and $cp_{min}$ is a user-defined parameter. $\epsilon_{0}$ depends on the initial school constraint violation \cite{takahama2005}:

\begin{equation}
\epsilon_0=\epsilon(0)=\frac{1}{2}\left(\frac{1}{N} \sum_{i=1}^{N} \phi_i(\textbf{x}) + min(\phi_i(\textbf{x}))\right).
\end{equation}

Individual movement operator in wrFSSe applies the $\epsilon$-comparison between the fishes' current and candidate positions. If candidate position is $>_\epsilon$ with relation to the current position, the movement is allowed.

\subsubsection{wrFSSg}

Gradient based individual movement operators were designed in order to guide the local search process in either phases of the search process intended to:

\begin{itemize}
\item Phase 1 - The gradient based individual movement is performed in order to allow the fishes to quickly achieve feasible regions;
\item Phase 2 - The gradient based individual movement is intended not to allow the fishes to escape the feasible regions.
\end{itemize}

In order to do so, the following steps are employed:

\begin{enumerate}
\item Calculate $\nabla \cdot \phi_i$;
\item Chose $K$ random directions and compute their director vectors $\textbf{u}_{ki}$;
\item Calculate the $K$ directional derivatives $\nabla_{\textbf{u}_{k} \phi_i}$ given by $(\nabla \cdot \phi_i) \cdot \textbf{u}_{ki}$;
\item Return $min\left( \nabla_{\textbf{u}_{ki}} \phi_i \right)$ in phase 1 or $min\left( | \nabla_{\textbf{u}_{ki}} \phi_i | \right)$ in phase 2.
\end{enumerate}

With this procedure, we intended to provide the individual movement operator with the directions containing high probabilities of improving constraint violation $\phi_i$ within phase 1 or improving fitness value maintaining fishes in feasible regions within phase 2 of the search process.

The gradient $(\nabla \cdot \phi_i)$ is calculated according to (\ref{gradient}) \cite{chootinan2006constraint}:

\begin{equation}
\label{gradient}
\nabla \cdot \phi_i = \frac{1}{e} \cdot \left[ \phi_i (\textbf{x}|x_j=x_j+e)-\phi_i(\textbf{x}) \text{, } \forall j=1,...,D \right],
\end{equation}
where $D$ is the problem dimensions number and $e$ is a pertubation constant.

The directional derivatives are computed by the inner product between the gradient and the director vectors.

The candidate position for the Individual movement operator is given by $step_{ind} \cdot min\left( \nabla_{\textbf{u}_{ki}} \phi_i\right)$ or $step_{ind} \cdot min\left( | \nabla_{\textbf{u}_{ki}} \phi_i | \right)$ for phases 1 and 2, respectively.

Gradient evaluation demands $D+1$ $\phi$ evaluations. Therefore, a user-defined probability $P_g$ is defined in order for each fish to check whether it will run the gradient based or the original Individual movement operator.

\subsubsection{wrFSSp}

A simple modification was proposed in wrFSS originating wrFSSp. A penalty approach was applied specifically in phase 2 in order to avoid that feasible fishes move to infeasible positions which improve their fitnesses. Thus, the objective function of phase 2 is defined as in (\ref{pfitness}):

\begin{equation}
\label{pfitness}
f_p(\textbf{x}) = f(\textbf{x}) + \phi(\textbf{x}).
\end{equation}

\section{Experiments}
\label{experiments}

In order to evaluate wrFSS different versions performance, a set of constrained optimization problems defined for CEC 2010 \cite{Mallipeddi2010} was solved.

The chosen problems are presented in Table \ref{problemsDef}  as well as their features. The problems selected to be included in the test set present different feasible regions. The feasible region is the ratio between the feasible portion and the whole search space.

\begin{table}[]
\centering
\caption{Chosen CEC 2010's Problems}
\label{problemsDef}
\begin{tabular}{ccccc}
\hline
\multirow{2}{*}{Problem} & \multirow{2}{*}{Search Space} & \multicolumn{2}{c}{Number of Constraints} & \multirow{2}{*}{\begin{tabular}[c]{@{}c@{}}Feasible\\ Region (10D)\end{tabular}} \\ \cline{3-4}
&                               & \textit{E}          & \textit{I}          &                                                                                  \\ \hline
C01                      & $[0;10]^{10}$                 & 0                   & 2                   & 0.997689                                                                         \\
C03                      & $[-1000;1000]^{10}$           & 1                   & 0                   & 0.000000                                                                         \\
C04                      & $[-50;50]^{10}$               & 4                   & 0                   & 0.000000                                                                         \\
C06                      & $[-600;600]^{10}$             & 2                   & 0                   & 0.000000                                                                         \\
C07                      & $[-140;140]^{10}$             & 0                   & 1                   & 0.505123                                                                         \\
C08                      & $[-140;140]^{10}$             & 0                   & 1                   & 0.379512                                                                         \\
C09                      & $[-500;500]^{10}$             & 1                   & 0                   & 0.000000                                                                         \\ \hline
\end{tabular}
\end{table}

Two levels of each parameter were chosen in order to evaluate which combination of them caused the best results for C01 and C03. Based on that, the best performers parameters sets on C01 were extended to C07 and C08. The parameters which generated best results to C03 were applied also in the tests with C04, C06 and C09 due to the similarities of the problems in terms of feasible region. Table \ref{parameters} presents the chosen parameters values used in the tests performed.

\begin{table}[]
\centering
\caption{Parameters Definition}
\label{parameters}
\begin{tabular}{ccccccccccccc}
\hline
\multirow{2}{*}{Problem} & \multicolumn{2}{c}{wrFSS}             & \multicolumn{4}{c}{wrFSSe}                                   & \multicolumn{4}{c}{wrFSSg}                             & \multicolumn{2}{c}{wrFSSp} \\ \cline{2-13} 
& $\sigma$ & $\tau$                     & $T_c$  & $cp_{min}$ & $\sigma$ & $\tau$                      & $P_g$  & $K$   & $\sigma$ & $\tau$                     & $\sigma$      & $\tau$     \\ \hline
C01                      & $5\%$    & \multicolumn{1}{c|}{$1\%$} & $60\%$ & $3$        & $5\%$    & \multicolumn{1}{c|}{$30\%$} & $10\%$ & $200$ & $50\%$   & \multicolumn{1}{c|}{$1\%$} & $5\%$         & $30\%$     \\
C03                      & $5\%$    & \multicolumn{1}{c|}{$1\%$} & $60\%$ & $8$        & $5\%$    & \multicolumn{1}{c|}{$30\%$} & $10\%$ & $50$  & $50\%$   & \multicolumn{1}{c|}{$1\%$} & $5\%$         & $30\%$     \\
C04                      & $5\%$    & \multicolumn{1}{c|}{$1\%$} & $60\%$ & $8$        & $5\%$    & \multicolumn{1}{c|}{$30\%$} & $10\%$ & $50$  & $50\%$   & \multicolumn{1}{c|}{$1\%$} & $5\%$         & $30\%$     \\
C06                      & $5\%$    & \multicolumn{1}{c|}{$1\%$} & $60\%$ & $8$        & $5\%$    & \multicolumn{1}{c|}{$30\%$} & $10\%$ & $50$  & $50\%$   & \multicolumn{1}{c|}{$1\%$} & $5\%$         & $30\%$     \\
C07                      & $5\%$    & \multicolumn{1}{c|}{$1\%$} & $60\%$ & $3$        & $5\%$    & \multicolumn{1}{c|}{$30\%$} & $10\%$ & $200$ & $50\%$   & \multicolumn{1}{c|}{$1\%$} & $5\%$         & $30\%$     \\
C08                      & $5\%$    & \multicolumn{1}{c|}{$1\%$} & $60\%$ & $3$        & $5\%$    & \multicolumn{1}{c|}{$30\%$} & $10\%$ & $200$ & $50\%$   & \multicolumn{1}{c|}{$1\%$} & $5\%$         & $30\%$     \\
C09                      & $5\%$    & \multicolumn{1}{c|}{$1\%$} & $60\%$ & $8$        & $5\%$    & \multicolumn{1}{c|}{$30\%$} & $10\%$ & $50$  & $50\%$   & \multicolumn{1}{c|}{$1\%$} & $5\%$         & $30\%$     \\ \hline
\end{tabular}
\end{table}

Table \ref{CECSolutions} presents the results obtained in 30 runs of each wrFSS variation when solving each of the aforementioned CEC's problem. In all the tests, the maximum number of iterations was set to $80000$ and the tolerance $\delta$ values $10^{-4}$ in all tests. All wrFSS variations include the Stagnation Avoidance Routine \cite{Monteiro2016} within the Individual movement operator. $\alpha$ was set to decay exponentially: $\alpha=0.8e^{-0.007*t}$, where $t$ is the current iteration.

\begin{table}[]
\centering
\caption{CEC 2010 Problems Solutions Results}
\label{CECSolutions}
\resizebox{\textwidth}{!}{
\begin{tabular}{cccccccccc}
& \textit{\textbf{}} & \multicolumn{2}{c}{\textbf{wrFSS}}      & \multicolumn{2}{c}{\textbf{wrFSSe}}     & \multicolumn{2}{c}{\textbf{wrFSSg}}     & \multicolumn{2}{c}{\textbf{wrFSSp}}     \\ \cline{3-10} 
\textit{\textbf{}}            &                    & \textit{Fitness} & \textit{Feasibility} & \textit{Fitness} & \textit{Feasibility} & \textit{Fitness} & \textit{Feasibility} & \textit{Fitness} & \textit{Feasibility} \\ \hline
\multirow{4}{*}{\textbf{C01}} & \textit{Mean}      & -5,91E-01        & 0,00E+00             & -4,03E-01        & 0,00E+00             & -5,76E-01        & 0,00E+00             & -6,93E-01        & 0,00E+00             \\
& \textit{SD}        & 4,83E-02         & 0,00E+00             & 1,17E-01         & 0,00E+00             & 3,16E-02         & 0,00E+00             & 1,64E-02         & 0,00E+00             \\
& \textit{min}       & -7,06E-01        & 0,00E+00             & -7,42E-01        & 0,00E+00             & -6,42E-01        & 0,00E+00             & -7,20E-01        & 0,00E+00             \\
& \textit{max}       & -4,95E-01        & 0,00E+00             & -2,67E-01        & 0,00E+00             & -5,15E-01        & 0,00E+00             & -6,63E-01        & 0,00E+00             \\ \hline
\multirow{4}{*}{\textbf{C03}} & \textit{Mean}      & 6,33E+12         & 4,45E-05             & 4,01E+09         & 1,55E-05             & 5,20E+13         & 5,11E-05             & 7,71E+12         & 3,54E-05             \\
& \textit{SD}        & 5,54E+12         & 6,55E-05             & 8,37E+09         & 4,02E-05             & 1,46E+14         & 7,14E-05             & 1,45E+13         & 6,03E-05             \\
& \textit{min}       & 6,82E+10         & 0,00E+00             & 1,40E+03         & 0,00E+00             & 1,35E+12         & 0,00E+00             & 1,56E+10         & 0,00E+00             \\
& \textit{max}       & 2,31E+13         & 1,77E-04             & 3,47E+10         & 1,24E-04             & 7,82E+14         & 2,25E-04             & 5,97E+13         & 1,50E-04             \\ \hline
\multirow{4}{*}{\textbf{C04}} & \textit{Mean}      & 2,23E+00         & 6,26E-04             & 5,60E+00         & 1,20E-03             & 1,88E+00         & 7,46E-04             & 1,55E+00         & 7,26E-04             \\
& \textit{SD}        & 5,37E+00         & 2,85E-04             & 7,16E+00         & 5,43E-04             & 4,64E+00         & 3,54E-04             & 4,24E+00         & 2,79E-04             \\
& \textit{min}       & 1,17E-02         & 2,03E-04             & 1,51E-02         & 2,32E-04             & 1,19E-02         & 0,00E+00             & 4,40E-03         & 1,02E-04             \\
& \textit{max}       & 1,62E+01         & 1,53E-03             & 1,62E+01         & 2,46E-03             & 1,62E+01         & 1,61E-03             & 1,40E+01         & 1,38E-03             \\ \hline
\multirow{4}{*}{\textbf{C06}} & \textit{Mean}      & 2,92E+02         & 0,00E+00             & -5,65E+02        & 0,00E+00             & -5,20E+00        & 0,00E+00             & 3,04E+02         & 0,00E+00             \\
& \textit{SD}        & 9,40E+01         & 0,00E+00             & 3,55E+00         & 0,00E+00             & 1,51E+02         & 0,00E+00             & 8,60E+01         & 0,00E+00             \\
& \textit{min}       & 4,86E+01         & 0,00E+00             & -5,71E+02        & 0,00E+00             & -4,52E+02        & 0,00E+00             & 1,12E+02         & 0,00E+00             \\
& \textit{max}       & 4,55E+02         & 0,00E+00             & -5,56E+02        & 0,00E+00             & 1,64E+02         & 0,00E+00             & 4,45E+02         & 0,00E+00             \\ \hline
\multirow{4}{*}{\textbf{C07}} & \textit{Mean}      & 5,09E+05         & 0,00E+00             & 5,01E+00         & 0,00E+00             & 5,88E+09         & 0,00E+00             & 4,32E+05         & 0,00E+00             \\
& \textit{SD}        & 3,17E+05         & 0,00E+00             & 6,63E+00         & 0,00E+00             & 4,23E+09         & 0,00E+00             & 2,40E+05         & 0,00E+00             \\
& \textit{min}       & 9,29E+04         & 0,00E+00             & 2,44E-01         & 0,00E+00             & 1,83E+09         & 0,00E+00             & 7,82E+04         & 0,00E+00             \\
& \textit{max}       & 1,74E+06         & 0,00E+00             & 3,44E+01         & 0,00E+00             & 2,02E+10         & 0,00E+00             & 1,17E+06         & 0,00E+00             \\ \hline
\multirow{4}{*}{\textbf{C08}} & \textit{Mean}      & 4,16E+09         & 0,00E+00             & 6,04E+01         & 0,00E+00             & 7,34E+09         & 0,00E+00             & 4,19E+09         & 0,00E+00             \\
& \textit{SD}        & 2,13E+09         & 0,00E+00             & 1,60E+01         & 0,00E+00             & 3,76E+09         & 0,00E+00             & 2,25E+09         & 0,00E+00             \\
& \textit{min}       & 4,64E+08         & 0,00E+00             & 3,72E+01         & 0,00E+00             & 1,01E+09         & 0,00E+00             & 1,11E+09         & 0,00E+00             \\
& \textit{max}       & 8,65E+09         & 0,00E+00             & 1,14E+02         & 0,00E+00             & 1,47E+10         & 0,00E+00             & 8,83E+09         & 0,00E+00             \\ \hline
\multirow{4}{*}{\textbf{C09}} & \textit{Mean}      & 4,57E+12         & 0,00E+00             & 3,61E+06         & 0,00E+00             & 9,52E+12         & 0,00E+00             & 4,39E+12         & 0,00E+00             \\
& \textit{SD}        & 2,06E+12         & 0,00E+00             & 1,40E+07         & 0,00E+00             & 4,89E+12         & 0,00E+00             & 1,79E+12         & 0,00E+00             \\
& \textit{min}       & 3,16E+11         & 0,00E+00             & 1,74E+03         & 0,00E+00             & 1,76E+12         & 0,00E+00             & 1,40E+12         & 0,00E+00             \\
& \textit{max}       & 7,84E+12         & 0,00E+00             & 6,66E+07         & 0,00E+00             & 2,59E+13         & 0,00E+00             & 7,97E+12         & 0,00E+00             \\ \hline
\end{tabular}}
\end{table}

From Table \ref{CECSolutions}, it is noticeable that all the proposed algorithms were able to reach feasible solutions in all runs for problems C01, C08 and C09, which are those containing relatively large feasible regions. The same did not happen in the cases of C03, C04, C06 and C09, heavily constrained problems due to equality constraints presence. Specifically in C04, only wrFSSg was able to find feasible individuals. In the other hand, in C06 and C09 all the approaches were able to find feasible individuals in all runs. In C03, all the variations of wrFSS were able to find feasible individuals, but not in al runs.

The difficult of wrFSS variations in order to tackle some heavily constrained problems is related to the search mechanisms employed. The Individual movement operator is based on a local search performed with a random jump. Therefore, in situations in which the feasible regions are very small, random jumps may neither guarantee that a fish can reach this region in phase 1 nor guarantee that a fish that has already reached it will keep there.

In the specific case of C03, for instance, the feasible region is composed by the line $x_i=x_1+1=...=x_D$. Thus, even when fishes are able to reach the line, becoming feasible and changing the search mode from phase 1 to phase 2, once they perform the Individual movement operator, the random jumps will not allow them to strictly move over the line. This drawback was tackled by the application of the $\epsilon-$method and the gradient based individual movement operators. However, these two methods still apply random jumps and then, depending on the topological features of the feasible regions, the algorithm could fail to exploit fitness in phase 2.

\begin{figure}[]
\centering
\subfigure[a][Best Fish Fitness Convergence in C03]{\includegraphics[width=5cm]{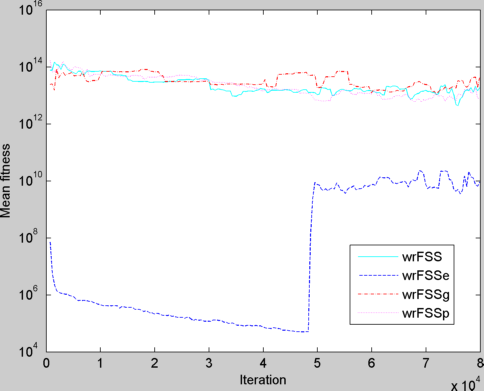}}
\qquad
\subfigure[b][Best Fish Feasibility Convergence in C03]{\includegraphics[width=5cm]{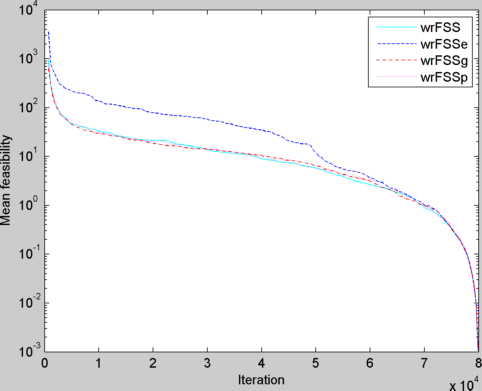}}

\subfigure[c][Best Fish Fitness Convergence in C09]{\includegraphics[width=5cm]{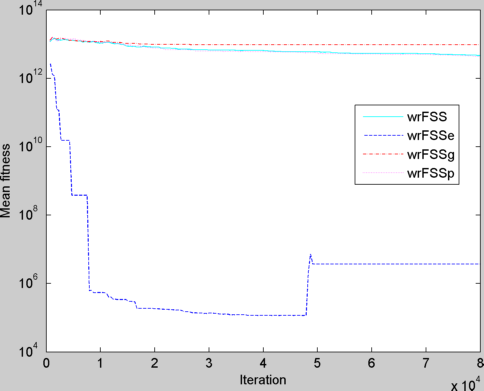}}
\qquad
\subfigure[d][Best Fish Feasibility Convergence in C09]{\includegraphics[width=5cm]{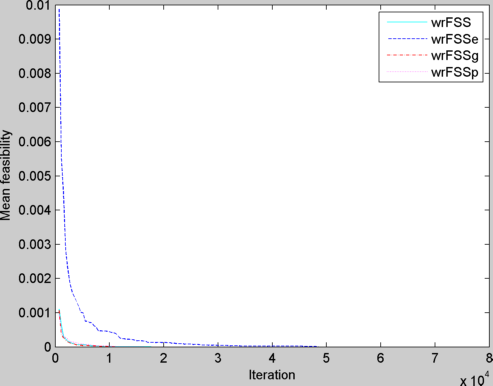}}

\caption{Best Fish Fitness and Feasibility Convergence}
\label{wrFSSConvergence}	
\end{figure}

Figure \ref{wrFSSConvergence} supports the aforementioned issues and displays the mean Best Fish's Fitness and Feasibility (constraint violation measure) along with the iterations. It is possible to notice that, in C03, most approaches are not able to improve fitness once the feasibility improves up to the end of the search process. This means that a long phase 1 happens in all versions, trying to find and keep the feasible line. The fast improving fitness in wrFSSe happens because of the characteristic relaxation of the constraint violation present in $\epsilon-$method. Moreover, in C09, It is possible to notice that all versions of wrFSS are able to reach feasible regions, but the fitness does not improve when that happens. Which means that once fishes reach feasible regions and change the search mode to phase 2, in few iterations the feasibility state degenerates due to random jumps in infeasible directions and phase 1 takes place again avoiding fitness convergence.

% \begin{figure}
% \subfloat[Best Fish Fitness Convergence in C03]{\label{C03Fitness}\includegraphics[width=0.46\textwidth, height=4cm]}}
% \qquad
% \subfloat[Best Fish Feasibility Convergence in C03]{\label{C03Feasibility}\includegraphics[width=0.46\textwidth, height=4cm]{Feasibility_C03.png}}\\
% \subfloat[Best Fish Fitness Convergence in C09]{\label{C09Fitness}\includegraphics[width=0.46\textwidth, height=4cm]{Fitness_C09.png}}
% \qquad
% \subfloat[Best Fish Feasibility Convergence in C09]{\label{C09Feasibility}\includegraphics[width=0.46\textwidth, height=4cm]{Feasibility_C09.png}}\\
% \centering
% \caption{Best Fish Fitness and Feasibility Convergence }
% \label{wrFSSConvergence}
% \end{figure}

A fitness comparison is provided in Table \ref{topTenComparison}. Three CEC 2010's Top Ten rated approaches namely $\epsilon$DEg \cite{Takahama2010}, E-ABC \cite{Mezura-Montes2010} and Co-CLPSO \cite{Liang2010} were selected to be compared with wrFSS variations.

The best mean fitness values reached among all the approaches as well as the best performing wrFSS variation are highlighted.

One can notice that $\epsilon$DEg outperforms all the other approaches in all the problems selected, except for C08 where Co-CLPSO reaches better results. However, regarding specifically wrFSS variations, it can be seen that wrFSSe is the best wrFSS variation in 5 out of 7 tests. Further, in 6 out of 7 tests performed, some wrFSS variation outperforms at least one of the three approaches selected for comparison.

\begin{table}[]
\centering
\caption{Fitness values comparison with other approaches}
\label{topTenComparison}
\resizebox{\textwidth}{!}{
\begin{tabular}{ccccccccc}
& \textit{\textbf{}} & \textbf{wrFSS} & \textbf{wrFSSe}    & \textbf{wrFSSg} & \textbf{wrFSSp}    & \textbf{$\epsilon$DEg}      & \textbf{Co-CLPSO}  & \textbf{E-ABC} \\ \hline
\multirow{2}{*}{\textbf{C01}} & \textit{Mean}      & -5,91E-01      & -4,03E-01          & -5,76E-01       & \textbf{-6,93E-01} & \textbf{-7,47E-01} & -7,34E-01          & -7,16E-01      \\
& \textit{SD}        & 4,83E-02       & 1,17E-01           & 3,16E-02        & 1,64E-02           & 1,32E-03           & 1,78E-02           & 2,69E-02       \\ \hline
\multirow{2}{*}{\textbf{C03}} & \textit{Mean}      & 6,33E+12       & \textbf{4,01E+09}  & 5,20E+13        & 7,71E+12           & \textbf{0,00E+00}  & 3,55E-01           & 2,45E+12       \\
& \textit{SD}        & 5,54E+12       & 8,37E+09           & 1,46E+14        & 1,45E+13           & 0,00E+00           & 1,78E+00           & 1,01E+12       \\ \hline
\multirow{2}{*}{\textbf{C04}} & \textit{Mean}      & 2,23E+00       & 5,60E+00           & 1,88E+00        & \textbf{1,55E+00}  & \textbf{-9,92E-06} & -9,34E-06          & 8,56E-01       \\
& \textit{SD}        & 5,37E+00       & 7,16E+00           & 4,64E+00        & 4,24E+00           & 1,55E-07           & 1,07E-06           & 3,01E+00       \\ \hline
\multirow{2}{*}{\textbf{C06}} & \textit{Mean}      & 2,92E+02       & \textbf{-5,65E+02} & -5,20E+00       & 3,04E+02           & \textbf{-5,79E+02} & \textbf{-5,79E+02} & 4,38E+02       \\
& \textit{SD}        & 9,40E+01       & 3,55E+00           & 1,51E+02        & 8,60E+01           & 3,63E-03           & 5,73E-04           & 8,60E+01       \\ \hline
\multirow{2}{*}{\textbf{C07}} & \textit{Mean}      & 5,09E+05       & \textbf{5,01E+00}  & 5,88E+09        & 4,32E+05           & \textbf{0,00E+00}  & 7,97E-01           & 7,16E+01       \\
& \textit{SD}        & 3,17E+05       & 6,63E+00           & 4,23E+09        & 2,40E+05           & 0,00E+00           & 1,63E+00           & 5,19E+01       \\ \hline
\multirow{2}{*}{\textbf{C08}} & \textit{Mean}      & 4,16E+09       & \textbf{6,04E+01}  & 7,34E+09        & 4,19E+09           & 6,73E+00           & \textbf{6,09E-01}  & 4,11E+02       \\
& \textit{SD}        & 2,13E+09       & 1,60E+01           & 3,76E+09        & 2,25E+09           & 5,56E+00           & 1,43E+00           & 9,36E+02       \\ \hline
\multirow{2}{*}{\textbf{C09}} & \textit{Mean}      & 4,57E+12       & \textbf{3,61E+06}  & 9,52E+12        & 4,39E+12           & \textbf{0,00E+00}  & 1,99E+10           & 2,02E+12       \\
& \textit{SD}        & 2,06E+12       & 1,40E+07           & 4,89E+12        & 1,79E+12           & 0,00E+00           & 9,97E+10           & 1,81E+12       \\ \hline
\end{tabular}}
\end{table}

\newpage
\section{Conclusion}
\label{conclusion}

Several problems within Industry an Academia are constrained. Therefore, many approaches try to employ metaheuristic procedures in order to efficiently solve the aforementioned class of problems. Different search strategies were developed and applied in either Evolutionary Computation and Swarm Intelligence techniques.

The first contribution in this work regards the proposal of a new approach in order to tackle constrained optimization tasks: the separation of objective function and constraint violation by the division of the search process in two phases. Phase 1 is intended to make the swarm to find many different feasible regions and, after that, phase 2 takes place in order to exploit the feasible regions in terms of fitness values.

This strategy, mainly in phase 1, requires a niching able algorithm. Thus, we selected wFSS, the multi-modal version of the Fish School Search algorithm, to be employed as base algorithm. Hence, we conceived a variation of wFSS named wrFSS embedding the division strategy. Moreover, we proposed three variations of wrFSS applying different strategies in order to improve its performance.

In order to evaluate the techniques proposed, seven problems from CEC 2010 were solved. Results show that wrFSS as well as its variations are able to solve many hard constrained optimization problems. However, in some cases, specifically in problems containing feasible regions presenting geometric conditions in which the widths in some directions are much higher than in others, the algorithm's local search procedure brings difficulties for wrFSS to keep solutions feasible once phase 1 finishes. Even so, in a comparison performed with three CEC 2010's approaches within top 10 winners has shown that some wrFSS variation outperforms one of these techniques in almost all solved problems in this work. wrFSSe was the best variation of the proposed versions.

It is important to highlight that the approaches used for comparison apply costly search mechanisms that were not employed in wrFSS. Local optimization procedures and external archive were applied in these techniques.

Based on the aforementioned, the proposed strategy of dividing the search process in two different phases and apply a niching swarm optimization technique in order to find many feasible regions in phase 1 is an interesting approach to be explored. In future works, improvements in wrFSS could include a hybridization between wrFSSe and wrFSSg in order to improve the local search ability of wrFSSe, the best performing wrFSS variation. Moreover, the inclusion of other resources such as local optimization operators as well as external archive related operators could be employed in order to solve the issues highlighted in this work.

\newpage

\bibliographystyle{abbrv}
\bibliography{bibliografia}

\begin{thebibliography}{10}

\bibitem{Akay2012}
B.~Akay and D.~Karaboga.
\newblock {Artificial bee colony algorithm for large-scale problems and
  engineering design optimization}.
\newblock {\em Journal of Intelligent Manufacturing}, 23(4):1001--1014, 2012.

\bibitem{Bastos2015}
C.~J.~A. Bastos-Filho and A.~C.~S. Guimar{\~{a}}es.
\newblock {Multi-Objective Fish School Search}.
\newblock {\em International Journal of Swarm Intelligence Research},
  6(1):23--40, 2015.

\bibitem{Bonyadi2013}
M.~Bonyadi, X.~Li, and Z.~Michalewicz.
\newblock {A hybrid particle swarm with velocity mutation for constraint
  optimization problems}.
\newblock {\em Proceeding of the fifteenth annual conference on Genetic and
  evolutionary computation conference - GECCO '13}, page~1, 2013.

\bibitem{Brajevic2013}
I.~Brajevic and M.~Tuba.
\newblock {An upgraded artificial bee colony (ABC) algorithm for constrained
  optimization problems}.
\newblock {\em Journal of Intelligent Manufacturing}, 24(4):729--740, 2013.

\bibitem{Brest2009}
J.~Brest.
\newblock {Constrained real-parameter optimization with
  $\epsilon$-self-adaptive differential evolution}.
\newblock {\em Studies in Computational Intelligence}, 198:73--93, 2009.

\bibitem{Campos2013}
M.~Campos and R.~A. Krohling.
\newblock {Hierarchical bare bones particle swarm for solving constrained
  optimization problems}.
\newblock {\em 2013 IEEE Congress on Evolutionary Computation, CEC 2013}, pages
  805--812, 2013.

\bibitem{chootinan2006constraint}
P.~Chootinan and A.~Chen.
\newblock Constraint handling in genetic algorithms using a gradient-based
  repair method.
\newblock {\em Computers \& operations research}, 33(8):2263--2281, 2006.

\bibitem{DeLimaNeto2013}
F.~B. {De Lima Neto} and M.~G.~P. {De Lacerda}.
\newblock {Multimodal fish school search algorithms based on local information
  for school splitting}.
\newblock {\em Proceedings - 1st BRICS Countries Congress on Computational
  Intelligence, BRICS-CCI 2013}, pages 158--165, 2013.

\bibitem{Deb2000}
K.~Deb.
\newblock {An efficient constraint handling method for genetic algorithms}.
\newblock {\em Computer Methods in Applied Mechanics and Engineering},
  186(2-4):311--338, 2000.

\bibitem{Filho2008}
C.~J. a.~B. Filho, F.~B. D.~L. Neto, A.~J. C.~C. Lins, A.~I.~S. Nascimento, and
  M.~P. Lima.
\newblock {A novel search algorithm based on fish school behavior}.
\newblock {\em Conference Proceedings - IEEE International Conference on
  Systems, Man and Cybernetics}, pages 2646--2651, 2008.

\bibitem{Hamza2015}
N.~Hamza, D.~Essam, and R.~Sarker.
\newblock {Constraint Consensus Mutation based Differential Evolution for
  Constrained Optimization}.
\newblock {\em IEEE Transactions on Evolutionary Computation}, (c):1--1, 2015.

\bibitem{Hu2002}
X.~Hu and R.~Eberhart.
\newblock {Solving Constrained Nonlinear Optimization Problems with Particle
  Swarm Optimization}.
\newblock {\em Optimization}, 2(1):1677--1681, 2002.

\bibitem{Jordehi2015}
A.~R. Jordehi.
\newblock {A review on constraint handling strategies in particle swarm
  optimisation}.
\newblock {\em Neural Computing and Applications}, 26(6):1265--1275, 2015.

\bibitem{Koziel1999}
S.~Koziel and Z.~Michalewicz.
\newblock {Evolutionary algorithms, homomorphous mappings, and constrained
  parameter optimization.}
\newblock {\em Evolutionary computation}, 7(1):19--44, 1999.

\bibitem{Becerra2016}
R.~{Landa Becerra} and C.~A.~C. Coello.
\newblock {Cultured differential evolution for constrained optimization}.
\newblock {\em Computer Methods in Applied Mechanics and Engineering},
  195(33-36):4303--4322, 2006.

\bibitem{Li2014}
X.~Li and M.~Yin.
\newblock {Self-adaptive constrained artificial bee colony for constrained
  numerical optimization}.
\newblock {\em Neural Computing and Applications}, 24(3-4):723--734, 2014.

\bibitem{Liang2010}
J.~J. Liang, S.~Zhigang, and L.~Zhihui.
\newblock {Coevolutionary comprehensive learning particle swarm optimizer}.
\newblock {\em 2010 IEEE World Congress on Computational Intelligence, WCCI
  2010 - 2010 IEEE Congress on Evolutionary Computation, CEC 2010},
  450001(2):1--8, 2010.

\bibitem{Lin2013}
C.-H. Lin.
\newblock {A rough penalty genetic algorithm for constrained optimization}.
\newblock {\em Information Sciences}, 241:119--137, 2013.

\bibitem{Liu2016}
J.~Liu, K.~L. Teo, X.~Wang, and C.~Wu.
\newblock {An exact penalty function-based differential search algorithm for
  constrained global optimization}.
\newblock {\em Soft Computing}, 20(4):1305--1313, 2016.

\bibitem{Madeiro2011}
S.~S. Madeiro, F.~B. {De Lima-Neto}, C.~J.~A. Bastos-Filho, and E.~M. {Do
  Nascimento Figueiredo}.
\newblock {Density as the segregation mechanism in fish school search for
  multimodal optimization problems}.
\newblock {\em Lecture Notes in Computer Science (including subseries Lecture
  Notes in Artificial Intelligence and Lecture Notes in Bioinformatics)}, 6729
  LNCS(PART 2):563--572, 2011.

\bibitem{Mallipeddi2010}
R.~Mallipeddi and P.~N. Suganthan.
\newblock {Ensemble of constraint handling techniques}.
\newblock {\em IEEE Transactions on Evolutionary Computation}, 14(4):561--579,
  2010.

\bibitem{Mezura2011}
E.~Mezura-Montes and C.~A. {Coello Coello}.
\newblock {Constraint-handling in nature-inspired numerical optimization: Past,
  present and future}.
\newblock {\em Swarm and Evolutionary Computation}, 1(4):173--194, 2011.

\bibitem{Mezura-Montes2010}
E.~Mezura-Montes and R.~E. Velez-Koeppel.
\newblock {Elitist Artificial Bee Colony for constrained real-parameter
  optimization}.
\newblock {\em 2010 IEEE World Congress on Computational Intelligence, WCCI
  2010 - 2010 IEEE Congress on Evolutionary Computation, CEC 2010}, 2010.

\bibitem{monteirob2016}
J.~B. Monteiro, I.~M.~C. Albuquerque, F.~B.~L. Neto, and F.~V.~S. Ferreira.
\newblock Comparison on novel fish school search approaches.
\newblock {\em 16th International Conference on Intelligent Systems Design and
  Applications}, 2016.

\bibitem{Monteiro2016}
J.~B. Monteiro, I.~M.~C. Albuquerque, F.~B.~L. Neto, and F.~V.~S. Ferreira.
\newblock Optimizing multi-plateau functions with {FSS-SAR} ({S}tagnation
  {A}voidance {R}outine).
\newblock {\em IEEE Symposium Series on Computational Intelligence}, 2016.

\bibitem{Sargo2014}
J.~A.~G. Sargo, S.~M. Vieira, J.~M.~C. Sousa, and C.~J. A.~B. Filho.
\newblock {Binary Fish School Search applied to feature selection: Application
  to ICU readmissions}.
\newblock {\em IEEE International Conference on Fuzzy Systems}, pages
  1366--1373, 2014.

\bibitem{takahama2005}
T.~Takahama and S.~Sakai.
\newblock {Contrained Optimization by $\epsilon$ Constrained Swarm Optimizer
  with $\epsilon$-level Control}.
\newblock In {\em 4th IEEE International Workshop on Soft Computing as
  Transdisciplinary Science and Technology}, pages 1019--1029, 2005.

\bibitem{takahama2006}
T.~Takahama and S.~Sakai.
\newblock {Constrained Optimization by the Constrained Differential Evolution
  with Gradient-Based Mutation and Feasible Elites}.
\newblock {\em IEEE Congress on Evolution Computation}, pages 1--8, 2006.

\bibitem{takahama2009}
T.~Takahama and S.~Sakai.
\newblock {Solving difficult constrained optimization problems by the
  $\epsilon$ constrained differential evolution with gradient-based mutation}.
\newblock {\em Studies in Computational Intelligence}, 198:51--72, 2009.

\bibitem{Takahama2010}
T.~Takahama and S.~Sakai.
\newblock {Constrained Optimization by the $\epsilon$ Constrained Differential
  Evolution with an archive and Gradient-Based Mutation}.
\newblock {\em IEEE Congress on Evolution Computation}, (1):1--8, 2010.

\bibitem{takahamab2005}
T.~Takahama, S.~Sakai, and N.~Iwane.
\newblock {Constrained Optimization by the $\epsilon$ Constrained Hybrid
  Algorithm of Particle Swarm Optimization and Genetic Algorithm}.
\newblock {\em Advances in Artificial Intelligence}, 3809(1):389--400, 2005.

\end{thebibliography}

\end{document}